\documentclass{article}
\usepackage{ijcai24}

\usepackage{times}
\usepackage{soul}
\usepackage{url}
\usepackage[hidelinks]{hyperref}
\usepackage[utf8]{inputenc}
\usepackage[small]{caption}
\usepackage{graphicx}
\usepackage{amsmath}
\usepackage{amsthm}
\usepackage{booktabs}
\usepackage{algorithm}
\usepackage{algorithmic}
\usepackage[switch]{lineno}
\usepackage{multirow}
\usepackage{amsfonts}

\title{PatchMixer: A Patch-Mixing Architecture for Long-Term Time Series Forecasting}

\author{
    Zeying Gong, Yujin Tang and Junwei Liang\thanks{Corresponding author}
    \affiliations
    Hong Kong University of Science and Technology (Guangzhou)
    \emails
    {zgong313@connect.hkust-gz.edu.cn}, {\{yujintang,junweiliang\}@hkust-gz.edu.cn}
}

\begin{document}

\maketitle

\begin{abstract}
Recently, transformers incorporating patch-based representations have set new benchmarks in long-term time series forecasting. This naturally raises an important question: Is the impressive performance of patch-based transformers primarily due to the use of patches rather than the transformer architecture itself? To explore this, we introduce \textbf{PatchMixer}, a patch-based CNN that enhances accuracy and efficiency through depthwise separable convolution.
Our experimental results on seven time-series forecasting benchmarks indicate that PatchMixer achieves relative improvements of  $3.9\%$, $11.6\%$, and $21.2\%$ in comparison to state-of-the-art Transformer, MLP, and CNN models, respectively. Additionally, it demonstrates \textbf{2-3} training and inference times faster than the most advanced method. We also found that optimizing the patch embedding parameters and enhancing the objective function enables PatchMixer to better adapt to different datasets, thereby improving the generalization of the patch-based approach. Code is available at: \href{https://github.com/Zeying-Gong/PatchMixer}{https://github.com/Zeying-Gong/PatchMixer}.

\end{abstract}

\section{Introduction}

Long-term time series forecasting (LTSF) remains a crucial challenge within machine learning, entailing the prediction of future data points based on historically observed sequences. LTSF applications across various domains include traffic flow estimation, energy management, and weather forecasting.

In the LTSF domain, prevalent methods are typically Transformer and MLP-based models, which have a global receptive field due to features like self-attention in Transformers and global operations in MLPs. Informer \cite{informer} and various Transformer variants \cite{autoformer,fedformer,pyraformer} have been developed for time series analysis. However, the simple MLP networks in \cite{dlinear} surprisingly outperformed previous models, prompting a shift to MLPs and a reevaluation of Transformers' ability to capture long-term temporal relationships. Recently, PatchTST \cite{patchtst}, a patch-based Transformer, demonstrated the continued dominance of Transformers in LTSF with superior predictive performance. This raises a key question: Is the impressive performance of patch-based Transformers due to the Transformer architecture itself or the patch-based input representation?

To answer this question, we might as well turn our focus to Convolutional Neural Networks (CNNs), which only have the local receptive field. Therefore, it is common to consider this architecture unsuitable for LTSF tasks. However, this contrast raises an intriguing possibility: achieving comparable or superior forecasting performance with patch presentation and convolutional core module could highlight the importance of data preprocessing techniques in the time series. 

In this paper, we introduce a novel patch-based model known as PatchMixer. It is based on a depthwise separable convolutional module and distinguished by its ``patch-mixing'' design, which (\romannumeral1) processes time series data in patches, retaining the sequential structure, and (\romannumeral2) captures dependencies within and across variables through shared weights, emphasizing the role of patch-based preprocessing in striking a balance between efficiency and performance. The main contributions are as follows:

\vspace{-5pt}
\begin{itemize}
  \item  PatchMixer surpasses the state-of-the-art (SOTA) Transformer, MLP, and CNN models in Mean Squared Error (MSE), achieving improvements of $3.9\%$, $11.6\%$, and $21.2\%$ respectively, across seven prominent long-term forecasting benchmarks. 
  \item  PatchMixer balances efficiency and performance, demonstrating a \textbf{3x} faster inference and \textbf{2x} faster training speed compared to the current SOTA models under the same configurations.
  \item  In our exploration of patch-based parameter optimization, we enhance the patch-mixing method's predictive performance and generalization capabilities by improving the patch embedding and the loss function.
\end{itemize}

\section{Related Work}

\label{sec::related_work}

\noindent \textbf{CNNs for long-term context.} Most temporal CNNs adopt complex architectures to extend their limited receptive fields. For example, the TCN model \cite{tcn} utilized dilated causal convolutions to expand the field. MICN \cite{micn} introduced a novel multi-scale hybrid decomposition and isometric convolution. TimesNet \cite{timesnet} applied a parallel convolutional structure in the Inception \cite{inceptionv1} style. Outside the realm of time series analysis, there are also a few CNN-based studies for long-term context \cite{s4,ckconv,hyena}. 

\textbf{Depthwise Separable Convolution.} This technique can be abbreviated as \textbf{DWConv}. It was first introduced in 2014 \cite{sifre2014rigid}, and widely adopted in computer vision. It gained prominence in the Inception V1 and V2 models \cite{inceptionv1,inceptionv2}, supported Google's MobileNet \cite{mobilenets} for mobile efficiency, and was scaled up in the Xception network \cite{xception}. ConvMixer \cite{convmixer} highlighted patch representation in computer vision task via it.

\section{The Patch-Mixing Design}

\begin{figure}[h]
  \centering
    \begin{center}
    \includegraphics[width=0.85\linewidth]{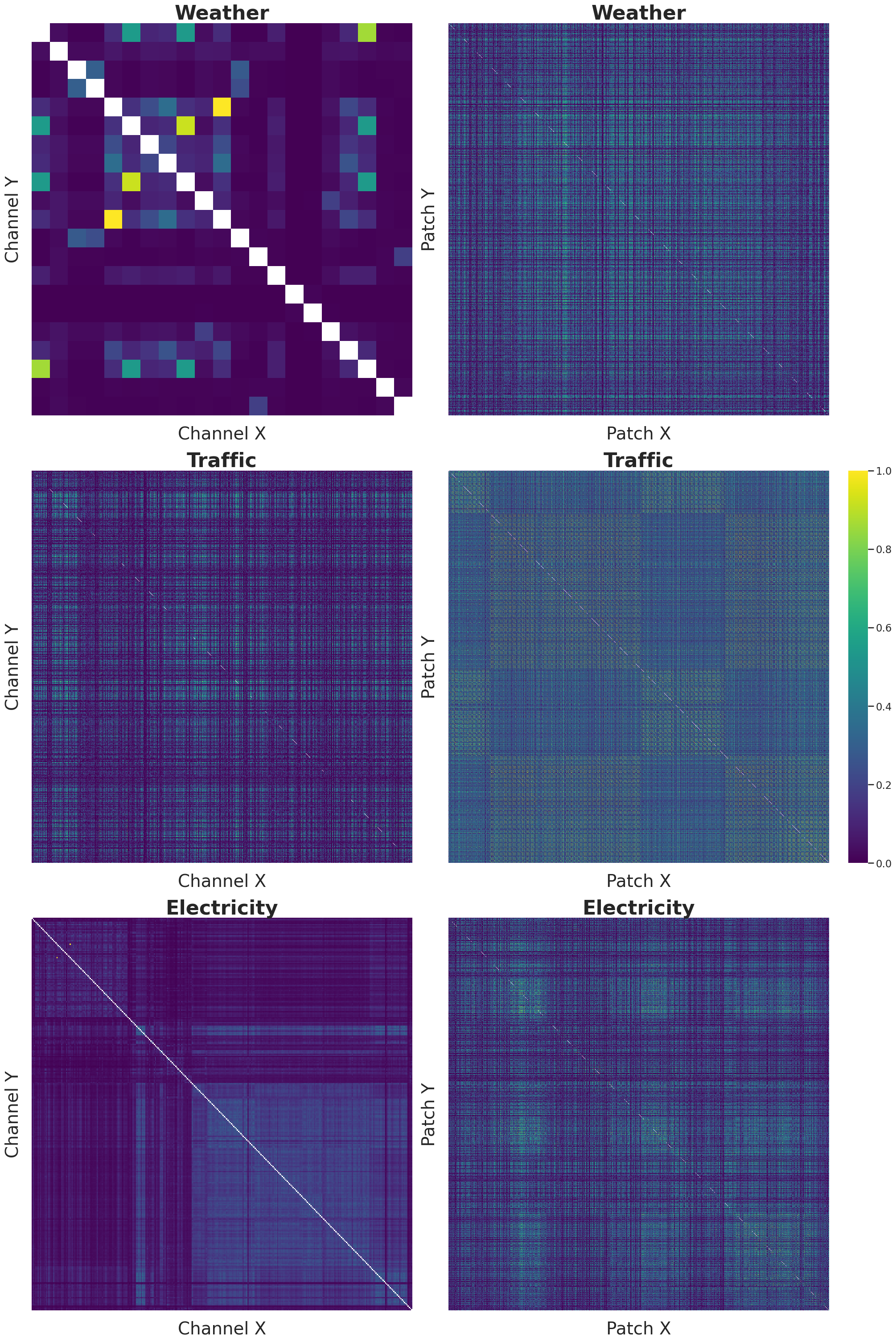}
    \end{center}
    \caption{Channel Dependency vs. Patch Dependency analysis was conducted on the 3 largest datasets: Traffic, Electricity, and Weather. The left panel shows the normalized mutual information among different variables, revealing sparse correlations. The right panel illustrates a pronounced intra-variable dependency within single-variable temporal patches, indicated by the more intense coloration.}
  \label{fig::combined_depend}
\end{figure}

In the LTSF field, we address the following task: Given a set of multivariate time series instances with a historical look-back window $L$: $X^{M \times L} = (x^{M}_1, \ldots, x^{M}_L)$, where each $x^{M}_t$ at time step $t$ represents a time series vector $x$ of $M$ variables. Our objective is to make predictions for the subsequent $T$ time steps, resulting in the prediction sequence $\hat{X}^{M \times T} = (x^{M}_{L+1}, \ldots, x^{M}_{L+T})$.

From a signal processing perspective, a multivariate time series is a signal with multiple channels. "Channel mixing" integrates information across variables, while "channel independence" treats each variable separately. This strategy was first used in \cite{dlinear} and further validated in \cite{patchtst}.

\begin{figure*}[t]
\begin{center}
\includegraphics[width=0.8\linewidth]{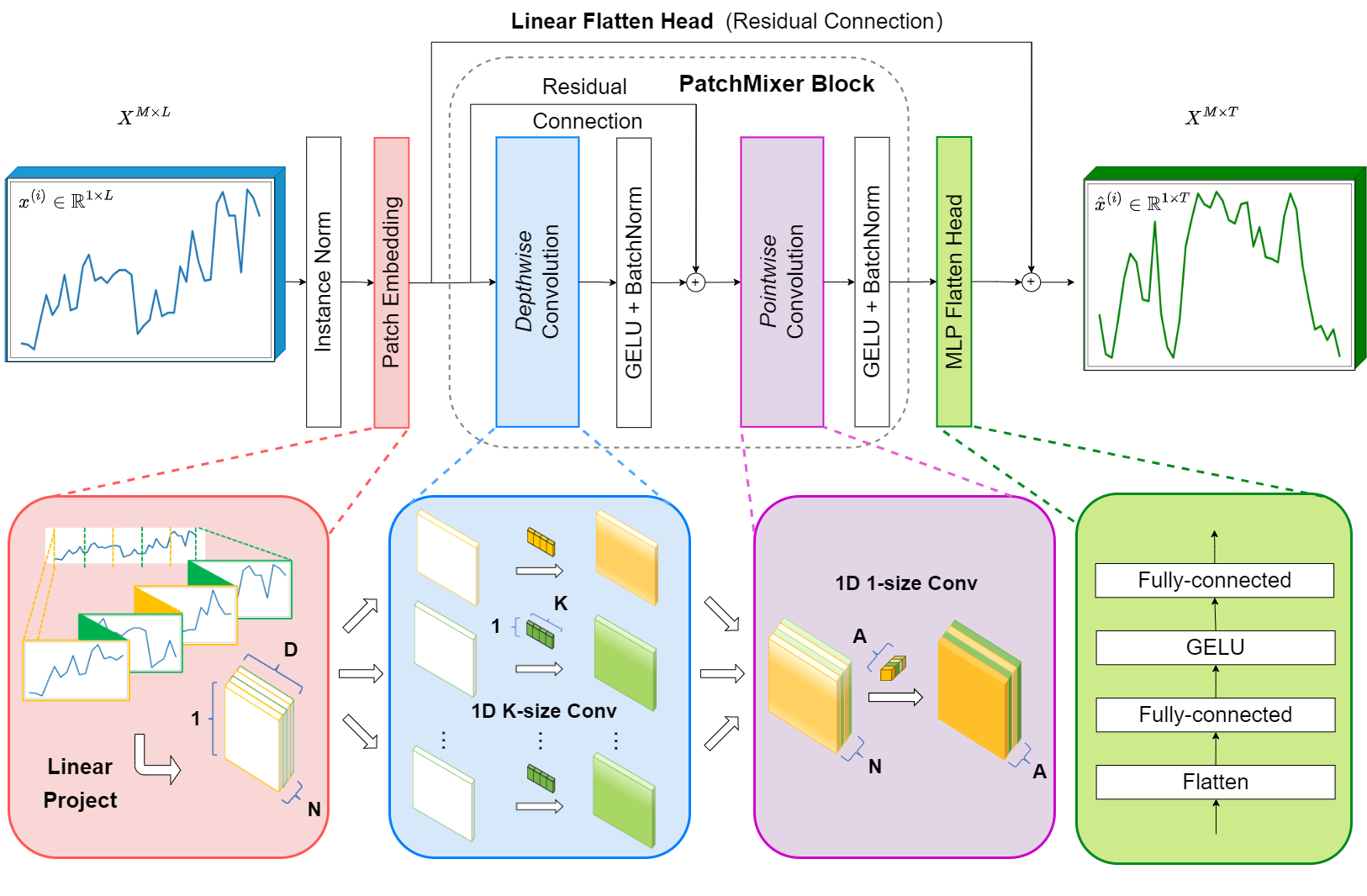}
\end{center}
\caption{PatchMixer overview.}
\label{fig::model_overview}
\end{figure*}

For the patch-based methods, a sliding window of size $P$ and stride $S$ unfolds across the extended series, generating $N$ patches of length $S$ (see Figure \ref{fig::model_overview}, left). This process starts by padding each univariate series, replicating the final value $S$ times to preserve edge information, thus $N = \lfloor \frac{(L-P)}{S} \rfloor + 2$.

Therefore, our PatchMixer abstracts the problem as follows. The multivariate time series $X^{M \times L} $ is split into $M$ univariate series $x^{(i)} \in \mathbb{R}^{1 \times L}$, where $i=1,..., M$. In patch embedding, they are transferred to temporal patches $x^{(i)}_p = (x^{(i)}_{1:P}, x^{(i)}_{1+S: P+S}, \ldots, x^{(i)}_{L-P+1:L})$, $P$ stands for patch length and $S$ for patch step. The model outputs a prediction for each series, denoted by $\hat{x}^{(i)} \in \mathbb{R}^{1 \times T}$ over the forecast horizon $T$. By aggregating these individual forecasts, we compose the final multivariate prediction $\hat{X}^{M \times T} = (x^{M}_{L+1}, \ldots, x^{M}_{L+T})$. 

We propose ``patch mixing'', which uses patching and channel independence to extract cross-patch temporal features. To validate our hypothesis, we measure the interdependence of channels and patches using \textbf{Normalized Mutual Information (NMI)}, which quantifies the shared information between two variables \cite{mi}, defined as:

\begin{equation}
NMI(X; Y) = \frac{2I(X; Y)}{H(X) + H(Y)}
\end{equation}
where $X$ and $Y$ are two random variables, $I(X; Y)$ is the mutual information, while $H(X)$ and $H(Y)$ are the respective entropies of $X$ and $Y$.

As Figure \ref{fig::combined_depend} shows, intra-variable temporal patterns hold significant mutual information in the main datasets. The improvement of patch mixing on prediction performance is evident from the experimental results in Table \ref{tab::ablation}.

\section{The PatchMixer Model}

\subsection{Model Structure}

The overall architecture of PatchMixer is depicted in Figure \ref{fig::model_overview}. We employ a patch-based DWConv module called PatchMixer Block. This design holds efficient parameter usage while maintaining a wide receptive field via patch embedding. Besides, we also devise dual forecasting heads, ensuring a holistic feature representation.

\subsection{PatchMixer Block}

PatchMixer Block utilizes DWConv as its core. It is composed of the following:

\noindent \textbf{Depthwise Convolution:} We utilize depthwise convolution with the kernel size $K=8$. 
Assuming that $x^{N \times D}$ represents one of the univariate series, $\sigma$ denotes an activation function GELU \cite{gelu}, $\text{BN}$ represents the BatchNorm operation and $l$ represents the current layer, this process is as follows:
\begin{small}
\begin{equation}
	x^{N \times D}_{l} = \text{\footnotesize{BN}}\left(\sigma(\text{Conv}_{_{N \to N}}(x^{N \times D}_{l-1}),{\text{kernel}=\text{step}=K})\right)
\label{eq_depthconv}
\end{equation}
\end{small}

\noindent \textbf{Pointwise Convolution:} It is applied to capture inter-patch feature correlations, enhanced by a residual connection.

\begin{small}
\begin{equation}
	x^{A \times D}_{l+1} = \text{\footnotesize{BN}}\left(\sigma(\text{\footnotesize{Conv}}_{_{N \to N}}(x^{N \times D}_{l}), {\text{kernel}=\text{step}=1})\right)
\label{eq_depthconv3}
\end{equation}
\end{small}
The above equations \ref{eq_depthconv3} demonstrate the process of the univariate series $x^{N \times D}$ in layer $l$ passing through the pointwise convolution kernel in layer $(l+1)$, while $\text{\footnotesize{Conv}}_{_{N \to N}}$ represents input and output channels are both $N$. 

\subsection{Dual Forecasting Heads}
PatchMixer introduces a dual-head mechanism. The residual connection spotlights linear trends, and the MLP head addresses nonlinear dynamics. The combination effectively captures a broad spectrum of temporal patterns in fine-to-coarse temporal modeling \cite{zhoujiaming}. 

\subsection{Instance Normalization}

This technique \cite{instance,revin} is applied before patching, with the original mean and standard deviation reintegrated post-forecasting to preserve the initial scale and distribution of the input time series.

\section{Experiments}

\subsection{Multivariate Long-term Forecasting}
\label{subsection::time series forecasting}

\textbf{Datasets.} We use $7$ popular multivariate datasets provided in \cite{autoformer} for forecasting. Detailed statistical data on the size of the datasets are as follows.

\begin{table}[h]
\centering
\scalebox{0.85}{
\begin{tabular}{c|cccc}
\toprule  
 & Variables & Timesteps & Frequencies \\
\midrule  
Weather & 21 & 52696 & 10 Minutes \\
Traffic & 862 & 17544 & 1 Hour \\
Electricity & 321 & 26304 & 1 Hour \\
ETTh1 & 7 & 17420 & 1 Hour \\
ETTh2 & 7 & 17420 & 1 Hour \\
ETTm1 & 7 & 69680 & 15 Minutes \\
ETTm2 & 7 & 69680 & 15 Minutes \\
\bottomrule 
\end{tabular}}
\label{tab::datasets}
\caption{Statistics of popular datasets used for benchmarking.}
\end{table}

\begin{itemize}
\item {Weather:} This dataset collects 21 meteorological indicators in Germany, such as temperature and humidity. 
\item{Traffic:} This dataset records the road occupancy rates from different sensors on San Francisco freeways. 
\item{Electricity:} This dataset describes 321 customers' hourly electricity consumption.
\item{ETT (Electricity Transformer Temperature):} It contains data from two electric transformers (1 and 2), each with two resolutions (15 minutes and 1 hour), resulting in 4 subsets: \textit{ETTm1}, \textit{ETTm2}, \textit{ETTh1}, and \textit{ETTh2}.
\end{itemize}

\noindent \textbf{Implementation details.} For a fair comparison, we adopt most of PatchTST's configurations. However, several differences are noteworthy: (\romannumeral1) The dimension of the Feed Forward Layer in PatchTST is $F = 256$ in most datasets. Our embedding dimension is set directly to $D=256$ across all datasets. (\romannumeral2) PatchTST has three layers of vanilla Transformer encoders, while our model achieves better performance with just one PatchMixer Block. (\romannumeral3) In our model's dual heads, the MLP head has two linear layers with GELU activation: one projects the hidden representation to $D = 2 \times T$ for the forecasting length $T$, and the other projects it to the final prediction target $D=T$. The linear head projects the embed vector directly from $N \times D$ to $T$. (\romannumeral4) Our model uses Mean Squared Error (MSE) and Mean Absolute Error (MAE) in a 1:1 ratio as the loss function. Details are in Section \ref{sec::ablation}.

\noindent \textbf{Baselines and metrics.} We choose the most representative LTSF models as our baselines, including Transformer-based models like PatchTST (\citeyear{patchtst}), FEDformer (\citeyear{fedformer}), Autoformer (\citeyear{autoformer}), Informer (\citeyear{informer}), in addition to two CNN-based models containing MICN (\citeyear{micn}) and TimesNet (\citeyear{timesnet}), with the significant MLP-based model DLinear (\citeyear{dlinear}). We employ commonly used evaluation metrics: MSE and MAE.   

\linespread{1.2}
\begin{table*}[t]
 \vspace{0.15in}
	\centering
	\begin{center}
        \begin{small}
        \scalebox{0.75}{
		\begin{tabular}{cc|c|cc|cc|cc|cc|cc|cc|cc|ccc}
			\cline{2-19}
&\multicolumn{2}{c|}{Models} & \multicolumn{2}{c}{\begin{tabular}[c]{@{}c@{}}\textbf{PatchMixer}\\ \textbf{(Ours)}\end{tabular}} & \multicolumn{2}{c}{\begin{tabular}[c]{@{}c@{}}PatchTST\\ \citeyear{patchtst}\end{tabular}} & \multicolumn{2}{c}{\begin{tabular}[c]{@{}c@{}}DLinear\\ \citeyear{dlinear}\end{tabular}} & \multicolumn{2}{c}{\begin{tabular}[c]{@{}c@{}}MICN\\ \citeyear{micn}\end{tabular}} & \multicolumn{2}{c}{\begin{tabular}[c]{@{}c@{}}TimesNet\\ \citeyear{timesnet}\end{tabular}} & \multicolumn{2}{c}{\begin{tabular}[c]{@{}c@{}}FEDformer\\ \citeyear{fedformer}\end{tabular}} & \multicolumn{2}{c}{\begin{tabular}[c]{@{}c@{}}Autoformer\\ \citeyear{autoformer}\end{tabular}} & \multicolumn{2}{c}{\begin{tabular}[c]{@{}c@{}}Informer\\ \citeyear{informer}\end{tabular}} \\ \cline{2-19}
&\multicolumn{2}{c|}{Metrics}                  &MSE&MAE&MSE&MAE&MSE&MAE&MSE&MAE&MSE&MAE&MSE&MAE&MSE&MAE&MSE&MAE\\
			\cline{2-19}
			&\multirow{4}*{\rotatebox{90}{Weather}}& 96    & \textbf{0.151} & \textbf{0.193} & \underline{0.152} & \underline{0.199}                                              & 0.176                                                    & 0.237  & 0.172 & 0.240                                                   & 0.165                                                    & 0.222                                                    & 0.238                                                     & 0.314                                                    & 0.249                                                     & 0.329                                                     & 0.354                                                    & 0.405                                                     \\
            &\multicolumn{1}{c|}{}& 192   & \textbf{0.194} & \textbf{0.236} & \underline{0.197} & \underline{0.243}                                                    & 0.220                                                    & 0.282 & 0.218                                                    & 0.281                                                         & 0.215                                                    & 0.264                                              & 0.275                                                     & 0.329                                                    & 0.325                                                     & 0.370                                                     & 0.419                                                    & 0.434                                                     \\
            &\multicolumn{1}{c|}{}& 336  & \textbf{0.225} & \textbf{0.267} & \underline{0.249} & \underline{0.283}                                                    & 0.265                                                    & 0.319 & 0.275                                                    & 0.329                                                      & 0.274                                                    & 0.304                                                 & 0.339                                                     & 0.377                                                    & 0.351                                                     & 0.391                                                     & 0.583                                                    & 0.543                                                    \\
            &\multicolumn{1}{c|}{}& 720   & \textbf{0.305} & \textbf{0.323} & \underline{0.320} & \underline{0.335} & 0.323                                                    & 0.362 & 0.314                                                    & 0.354                                                     & 0.339                                                    & 0.349                                                  & 0.389                                                     & 0.409                                                    & 0.415                                                     & 0.426                                                     & 0.916                                                    & 0.705                                                   \\
			\cline{2-19}
			&\multirow{4}*{\rotatebox{90}{Traffic}}& 96  & \textbf{0.363} & \textbf{0.245} & \underline{0.367} & \underline{0.251}  & 0.410                                                    & 0.282 & 0.479                                                    & 0.295                                                   & 0.593                                                    & 0.321                                                                   & 0.576                                                     & 0.359                                                    & 0.597                                                     & 0.371                                                     & 0.733                                                    & 0.410                                                     \\
            &\multicolumn{1}{c|}{} & 192  & \textbf{0.384} & \textbf{0.254} & \underline{0.385} & \underline{0.259} & 0.423                                                    & 0.287                                                   & 0.482 & 0.297 & 0.617                                                    & 0.336                                                    & 0.610                                                     & 0.380                                                    & 0.607                                                     & 0.382                                                     & 0.777                                                    & 0.435                                                    \\
            &\multicolumn{1}{c|}{}& 336   & \textbf{0.393} & \textbf{0.258} & \underline{0.398} & \underline{0.265} & 0.436                                                    & 0.296                                                   & 0.492 & 0.297 & 0.629                                                    & 0.336                                                    & 0.608                                                     & 0.375                                                    & 0.623                                                     & 0.387                                                     & 0.776                                                    & 0.434                                                    \\
            &\multicolumn{1}{c|}{}& 720  & \textbf{0.429} & \textbf{0.283} & \underline{0.434} & \underline{0.287} & 0.466                                                    & 0.315                                                   & 0.510 & 0.309 & 0.640                                                    & 0.350                                                    & 0.621                                                     & 0.375                                                    & 0.639                                                     & 0.395                                                     & 0.827                                                    & 0.466                                                     \\
            \cline{2-19}
			&\multirow{4}*{\rotatebox{90}{Electricity}}& 96   & \textbf{0.129} & \textbf{0.221} & \underline{0.130} & \underline{0.222} & 0.140                                                    & 0.237
   & 0.153                                                    & 0.264 & 0.168                                                    & 0.272                                                      & 0.186                                                     & 0.302                                                    & 0.196                                                     & 0.313                                                     & 0.304                                                    & 0.393                                                   \\
			&\multicolumn{1}{c|}{}& 192 & \textbf{0.144} & \textbf{0.237} & \underline{0.148} & \underline{0.240} & 0.153                                                    & 0.249
   & 0.175                                                    & 0.286 & 0.184                                                    & 0.289                                                    & 0.197                                                     & 0.311                                                    & 0.211                                                     & 0.324                                                     & 0.327                                                    & 0.417                                                    \\
			&\multicolumn{1}{c|}{}& 336  & \textbf{0.164} & \textbf{0.257} & \underline{0.167} & \underline{0.261} & 0.169                                                    & 0.267           & 0.192                                                    & 0.303                                                 & 0.198                                                    & 0.300                                                    & 0.213                                                     & 0.328                                                    & 0.214                                                     & 0.327                                                     & 0.333                                                    & 0.422                                                    \\
			&\multicolumn{1}{c|}{}& 720 & \textbf{0.200} & \textbf{0.289} & \underline{0.202} & \underline{0.291} & 0.203                                                    & 0.301 & 0.215                                                    & 0.323                                                   & 0.220                                                    & 0.320                                                    & 0.233                                                     & 0.344                                                    & 0.236                                                     & 0.342                                                     & 0.351                                                    & 0.427                                                    \\
			\cline{2-19}
			&\multirow{4}*{\rotatebox{90}{ETTh1}}& 96  & \textbf{0.353} & \textbf{0.381} & \underline{0.375} & \underline{0.399} &  \underline{0.375}                                                    & \underline{0.399} & 0.405                                                    & 0.430                                                   & 0.384                                                    & 0.402                                                    & 0.376                                                     & 0.415                                                    & 0.435                                                     & 0.446                                                     & 0.941                                                    & 0.769                                                    \\
            &\multicolumn{1}{c|}{}& 192  & \textbf{0.373} & \textbf{0.394} & 0.414 & 0.421  & \underline{0.405}                                                    & \underline{0.416}    & 0.447                                                    & 0.468                                                & 0.436                                                    & 0.429                                                    & 0.423                                                     & 0.446                                                    & 0.456                                                     & 0.457                                                     & 1.007                                                    & 0.786                                                 \\
            &\multicolumn{1}{c|}{}& 336 & \textbf{0.392} & \textbf{0.414} & \underline{0.431} & \underline{0.436} & 0.439                                                    & 0.443  & 0.579                                                    & 0.549                                                  & 0.491                                                    & 0.469                                                    & 0.444                                                     & 0.462                                                    & 0.486                                                     & 0.487                                                     & 1.038                                                    & 0.784                                                    \\
            &\multicolumn{1}{c|}{}& 720   & \textbf{0.445} & \textbf{0.463} & \underline{0.449} & \underline{0.466}                            & 0.472                                                    & 0.490   & 0.699                                                    & 0.635                                                & 0.521                                                    & 0.500                                                    & 0.469                                                     & 0.492                                                    & 0.515                                                     & 0.517                                                     & 1.144                                                    & 0.857                                                    \\
			\cline{2-19}
			&\multirow{4}*{\rotatebox{90}{ETTh2}}& 96 & \textbf{0.225} & \textbf{0.300} & \underline{0.274} & \underline{0.336}& 0.289                                                    & 0.353
   & 0.349                                                    & 0.401 & 0.340                                                    & 0.374                                                    & 0.332                                                     & 0.374                                                    & 0.332                                                     & 0.368                                                     & 1.549                                                    & 0.952                                                   \\
            &\multicolumn{1}{c|}{}& 192  & \textbf{0.274} & \textbf{0.334} & \underline{0.339} & \underline{0.379}  & 0.383                                                    & 0.418   & 0.442                                                    & 0.448                                                & 0.402                                                    & 0.414                                                    & 0.407                                                     & 0.446                                                    & 0.426                                                     & 0.434                                                     & 3.792                                                    & 1.542                                                    \\
            &\multicolumn{1}{c|}{}& 336 & \textbf{0.317} & \textbf{0.368} & \underline{0.331} & \underline{0.380}  & 0.480                                                    & 0.465  & 0.652                                                    & 0.569                                                  & 0.452                                                    & 0.452                                                    & 0.400                                                     & 0.471                                                    & 0.477                                                     & 0.479                                                     & 4.215                                                    & 1.642                                                   \\
            &\multicolumn{1}{c|}{}& 720 & \underline{0.393} & \underline{0.426} & \textbf{0.379} & \textbf{0.422} & 0.605                                                    & 0.551  & 0.800                                                   & 0.652                                                   & 0.462                                                    & 0.468                                                    & 0.412                                                     & 0.469                                                    & 0.453                                                     & 0.490                                                     & 3.656                                                    & 1.619                                                     \\
			\cline{2-19}
			&\multirow{4}*{\rotatebox{90}{ETTm1}}& 96 & \underline{0.291} & \textbf{0.340} & \textbf{0.290} & \underline{0.342} & 0.299                                                    & 0.343   & 0.302                                                    & 0.352                                                 & 0.340                                                    & 0.377                                                    & 0.326                                                     & 0.390                                                    & 0.505                                                     & 0.475                                                     & 0.626                                                    & 0.560                                                     \\
            &\multicolumn{1}{c|}{}& 192 & \textbf{0.325} & \textbf{0.362} & \underline{0.332} & 0.369 & 0.335                                                    & \underline{0.365} & 0.342                                                    & 0.380                                                   & 0.374                                                    & 0.387                                                    & 0.365                                                     & 0.415                                                    & 0.553                                                     & 0.496                                                     & 0.725                                                    & 0.619                                                     \\
            &\multicolumn{1}{c|}{}& 336  & \textbf{0.353} & \textbf{0.382} & \underline{0.366}                                                     &  0.453                                                   & 0.369                                                    & \underline{0.386}  & 0.381                                                    & 0.403                                                  & 0.392                                                    & 0.413                                                    & 0.392                                                     & 0.425                                                    & 0.621                                                     & 0.537                                                     & 1.005                                                    & 0.741                                                   \\
            &\multicolumn{1}{c|}{}& 720  & \textbf{0.413} & \textbf{0.413} & \underline{0.420}                                                     & 0.533                                                   & 0.425                                                    & \underline{0.421} & 0.434                                                    & 0.447                                                  & 0.433                                                    & 0.436                                                    & 0.446                                                     & 0.458                                                    & 0.671                                                     & 0.561                                                     & 1.133                                                    & 0.845                                                  \\
			\cline{2-19}
			&\multirow{4}*{\rotatebox{90}{ETTm2}} & 96   & 0.174 & \underline{0.256} & \textbf{0.165} & \textbf{0.255}  & \underline{0.167}                                                    & 0.260  & 0.188                                                    & 0.286                                                  & 0.183                                                    & 0.271                                                    & 0.180                                                     & 0.271                                                    & 0.255                                                     & 0.339                                                     & 0.355                                                    & 0.462                                                    \\
            &\multicolumn{1}{c|}{}& 192  & 0.227 & \underline{0.295} & \textbf{0.220} & \textbf{0.292}  & \underline{0.224}                                                    & 0.303 & 0.236                                                    & 0.320                                                  & 0.242                                                    & 0.309                                                    & 0.252                                                     & 0.318                                                    & 0.281                                                     & 0.340                                                     & 0.595                                                    & 0.586                                                    \\
            &\multicolumn{1}{c|}{}& 336   & \textbf{0.266} & \textbf{0.323} & \underline{0.278} & \underline{0.329} & 0.281                                                    & 0.342  & 0.295                                                    & 0.355                                                  & 0.304                                                    & 0.348                                                    & 0.324                                                     & 0.364                                                    & 0.339                                                     & 0.372                                                     & 1.270                                                    & 0.871                                                  \\
            &\multicolumn{1}{c|}{}& 720 & \textbf{0.344} & \textbf{0.372} & \underline{0.367} & \underline{0.385}  & 0.397                                                    & 0.421 & 0.422                                                    & 0.445                                                    & 0.385                                                    & 0.400                                                    & 0.410                                                     & 0.420                                                    & 0.433                                                     & 0.432                                                     & 3.001                                                    & 1.267                                                    \\
			\cline{2-19}
		\end{tabular}
	}
\linespread{1}
\end{small}
\end{center}
\caption{Multivariate long-term forecasting results with our model PatchMixer. The prediction lengths $T\in \{96, 192, 336, 720\}$ for all datasets. We report $L = 336$ for PatchMixer and PatchTST, and
the best result in $L = 24, 48, 96, 192, 336$ for the other baseline models by default. Thus it could be a strong baseline. The best results are in \textbf{bold} and the second best results are in \underline{underlined}.}
\label{tab::multivariate}
\end{table*}

\noindent \textbf{Results.} Table \ref{tab::multivariate} presents the multivariate long-term forecasting results. Our model outperforms all baselines on the three largest benchmarks: Traffic, Electricity, and Weather. On other datasets, it achieves the best performance for most prediction lengths. Quantitatively, PatchMixer shows improvements over the SOTA transformer PatchTST with reductions of $\mathbf{3.9\%}$ in MSE and $\mathbf{3.0\%}$ in MAE. Compared to the leading MLP model DLinear, it reduces MSE by $\mathbf{11.6\%}$ and MAE by $\mathbf{9.4\%}$ and achieves reductions of $\mathbf{21.2\%}$ in MSE and $\mathbf{12.5\%}$ in MAE over the most advanced CNN model TimesNet.

\subsection{Ablation Study}
\label{sec::ablation}

\textbf{Ablation of PatchMixer Block.} We conducted an ablation study by deleting or substituting the corresponding part, including patching, DWConv, and dual heads, with identical configurations for each variant.

As indicated in Table \ref{tab::ablation}, our findings highlight several key observations: (\romannumeral1) The patch technique enhances performance across various dataset sizes. (\romannumeral2) The convolutional module outperforms the attention mechanism across all datasets. (\romannumeral3) The linear head is more effective on small datasets, while the MLP head excels on larger ones; the dual-head combination provides balanced performance. (\romannumeral4) Position encoding tends to reduce accuracy in patch-based models for LTSF tasks, with better performance observed without it in two out of three datasets.

\begin{table}[h]
\centering
\scalebox{0.75}{
\begin{tabular}{c|cccc}
\toprule  
 \multirow{2}{*}{Dataset} & Electricity & Weather & ETTh2 \\
 & (Large) & (Medium) &  (Small) \\
\midrule  
Patch + DWConv + Dual Heads  & \underline{0.200} & \textbf{0.305} & \underline{0.393} \\
DWConv + Dual Heads & 0.211 & 0.322 & 0.415 \\
Patch + Linear Head & 0.210 & 0.329 & \textbf{0.383} \\
Patch + MLP Head  & \textbf{0.199} & 0.321 & 0.403 \\
Patch + Dual Heads & \underline{0.200} & 0.320 & 0.400 \\
Patch + Attention + Dual Heads & 0.203 & 0.319 & 0.396 \\
Patch + Attention + Dual Heads (wo. pos) & 0.202 & 0.320 & 0.395 \\
\bottomrule 
\end{tabular}}
\caption{Ablation Study across diverse dataset sizes, under $L=336$ and $T=720$ on each dataset. MSE evaluates the results, while the best are in \textbf{bold} and the second best results are in \underline{underlined}.}
\label{tab::ablation}
\end{table}

\begin{figure}[h]
\begin{center}
\includegraphics[width=\linewidth]{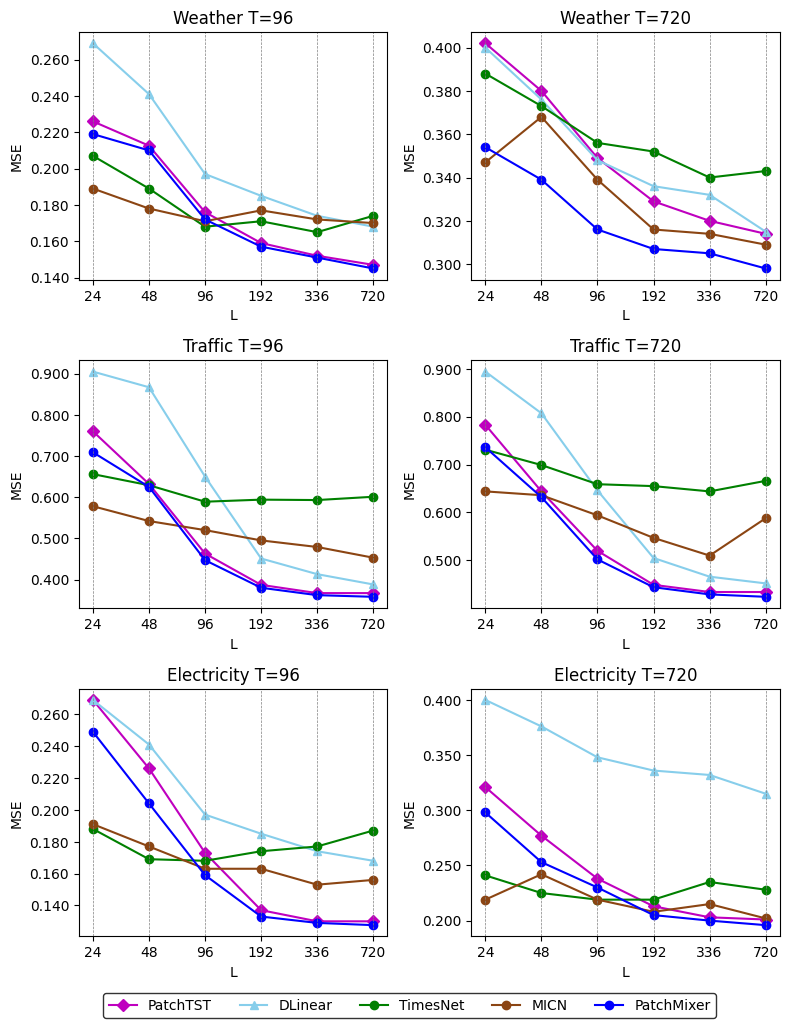}
\end{center}
\caption[MSE scores with varying look-back windows on top 3 largest datasets]{MSE scores with varying look-back windows on top 3 largest datasets. We report the top 5 methods for better observation. The look-back windows $L=[24,48,96,192,336,720]$, and the prediction horizons $T=[96, 720]$. }
\label{fig::varying_lookback}
\end{figure}

\noindent \textbf{Varying Look-back Windows.} Figure \ref{fig::varying_lookback} shows the impact of historical length on forecasting accuracy. Key trends are: (\romannumeral1) Traditional CNN-based methods show fluctuating performance with increasing input length, while PatchMixer's loss decreases steadily, utilizing longer look-back windows effectively. (\romannumeral2) Across varying input lengths, PatchMixer consistently outperforms PatchTST, showing the patch-based approach, rather than the Transformer architecture, plays a key role in enhancing performance.

\subsection{Efficiency Analysis}

\noindent \textbf{Comparison of Model Complexity} We analyze the computational complexity of PatchMixer, contrasting it with the previous SOTA method PatchTST using the same configuration, with the number of patches $N$ and the embedding dimension $D$. The convolutional approach has a kernel size of $K$, with $D>>N>K$. Table \ref{tab::complexity} shows the complexity results and \textbf{multiply-accumulate operations (MACs)} for this configuration. For comparison, we also include the computational cost of a standard convolution to emphasize the significant computational savings mainly due to the DWConv module.

\linespread{1.2}
\begin{table}[ht]
\centering
\begin{small}
	\scalebox{0.85}{
\begin{tabular}{c|c|c}
\hline
\textbf{Network Type}                     & \textbf{Computational Complexity}        & \textbf{MACs}              \\ \hline
Attention Mechanism                         & \( O(N \cdot D^2 + N^2 \cdot D) \)    & 293.63M               \\ \hline
Standard Convolution                   & \( O(N^2 \cdot D \cdot K) \)           & 175.57M              \\ \hline
PatchMixer Block           & \( O(N^2 \cdot D + N \cdot D \cdot K) \)     & 66.32M    \\ \hline
\end{tabular}
} 
\end{small}
\caption{Comparison of efficiency between patch-based Transformer and CNN models under $L=336$ and $T=720$ on the ETTm1 dataset. Note that $D >> N > K$.}
\label{tab::complexity}
\end{table}
\linespread{1}

\begin{figure}[h]
\begin{center}
\includegraphics[width=0.9\linewidth]{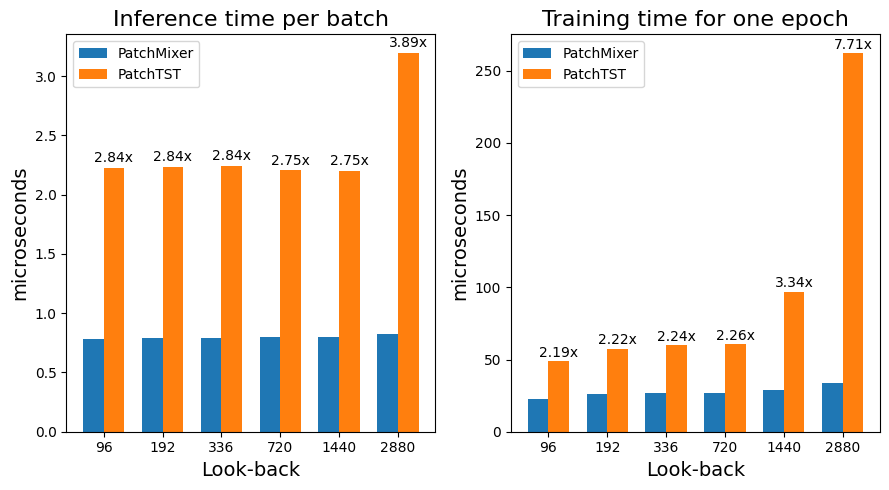}
\end{center}
\caption{PatchMixer vs. PatchTST: Comparison of Efficiencies.}
\label{fig::speed}
\end{figure}

\noindent \textbf{Training and Inference Efficiency.} As shown in Figure \ref{fig::speed}, we conducted experiments on the ETTm1 dataset using a batch size of 8 across 7 variables with identical configurations. We evaluated look-back lengths from 96 to 2880. 

Our results show two key improvements: PatchMixer is 3 times faster in inference and twice as fast in training as PatchTST. Additionally, PatchTST's performance is sensitive to look-back length, while PatchMixer shows fewer fluctuations with increasing input length, demonstrating its stability.

\subsection{Patch Embedding and Loss Optimization }

We further explore the parametric properties of the patch-mixing architecture. The predictive performance of patch-based methods can be significantly improved by optimizing the patch embedding and enhancing the objective function.

\begin{figure}[h]
\begin{center}
\includegraphics[width=\linewidth]{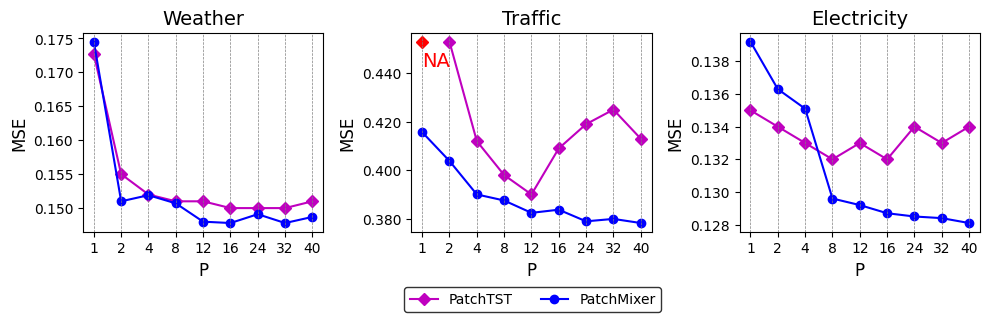}
\end{center}
\caption[MSE scores with varying patch length on the top $3$ largest datasets]{MSE scores with varying patch length on the top $3$ largest datasets. The patch length $P=[1,2,4,8,12,16,24,32,40]$, where $L=336$ and $T=96$. ``NA'' means the setup runs out of GPU memory (NVIDIA GTX4090 24GB) even with batch size 1.}
\label{fig::varying_patch}
\end{figure}

\noindent \textbf{Varying Patch Length.} Figure \ref{fig::varying_patch} illustrates how patch lengths affect performance. For patch-based methods, losses generally decrease or show minor fluctuations with larger patches. Notably, PatchMixer consistently achieves greater performance gains at larger patch sizes.

\begin{figure}[h]
\begin{center}
\includegraphics[width=\linewidth]{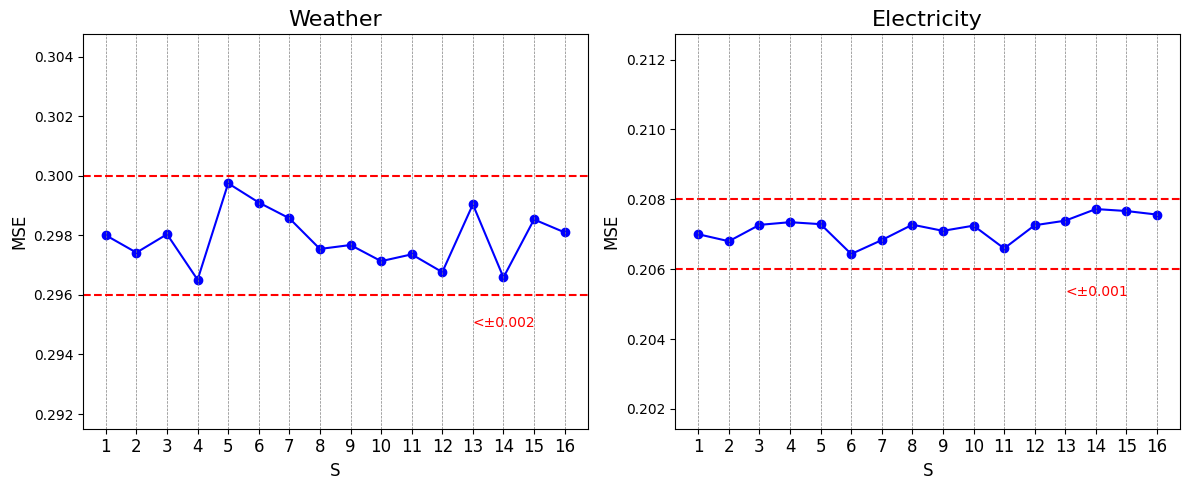}
\end{center}
\caption{MSE scores with varying patch strides $S$ from $1$ to $16$ where the look-back window is 336 and the prediction length is 720. }
\label{fig::vary_overlaps}
\end{figure}

\noindent \textbf{Varying Patch Strides.} In Figure \ref{fig::vary_overlaps}, as the patch stride increases, the MSE oscillates irregularly within a range of ±0.002. Besides, the lowest loss function points differ across datasets, suggesting that the patch stride parameter doesn't significantly aid in optimization.

\noindent \textbf{Loss Function.} We study the effects of different loss functions in Table \ref{tab::loss}. LTSF tasks have mainly used MSE, while a few models like \cite{scinet} utilized MAE. Recent research has introduced SmoothL1Loss \cite{petformer}, which combines aspects of both MSE and MAE. This observation leads us to explore a better loss approach.

\linespread{1.2}
\begin{table*}[t]
\vspace{0.15in}
	\centering
        \begin{small}
	\scalebox{0.75}{
		\begin{tabular}{cc|c|cc|cc|cc|cc|cc|cc|cc|ccc}
		    \cline{2-19}
			&\multicolumn{2}{c|}{\multirow{2}{*}{Models}}& \multicolumn{8}{c|}{PatchMixer}&  \multicolumn{8}{c}{PatchTST}& \\
			\cline{4-19}
			&\multicolumn{2}{c|}{}& \multicolumn{2}{c|}{MSE+MAE}& \multicolumn{2}{c|}{MSE}& \multicolumn{2}{c|}{MAE}& \multicolumn{2}{c|}{SmoothL1loss} & \multicolumn{2}{c|}{MSE+MAE}& \multicolumn{2}{c|}{MSE}& \multicolumn{2}{c|}{MAE}& \multicolumn{2}{c}{SmoothL1loss}&\\
			\cline{2-19}
			&\multicolumn{2}{c|}{Metric}&MSE&MAE&MSE&MAE&MSE&MAE&MSE&MAE&MSE&MAE&MSE&MAE&MSE&MAE&MSE&MAE\\
			\cline{2-19}
			&\multirow{4}*{\rotatebox{90}{Weather}}& 96    & \textbf{0.149} & 0.193 & 0.154 & 0.196 & 0.152 & 0.190 & 0.151 & 0.193 & 0.151 & 0.192 & 0.152 & 0.199 & 0.152 & 0.186  & 0.150 & 0.191\\
            &\multicolumn{1}{c|}{}& 192   & \textbf{0.191} & 0.233 & 0.197 & 0.237 & 0.193 & 0.231 & 0.194 & 0.237 & 0.195 & 0.234 & 0.197 & 0.243  & 0.197 & \textbf{0.230} & 0.196 & 0.236  \\
            &\multicolumn{1}{c|}{}& 336   & \textbf{0.225} & 0.269 & \textbf{0.225} & 0.267 & 0.227 & \textbf{0.265} & 0.229 & 0.272 & 0.247 & 0.275 & 0.249 & 0.283& 0.250 & 0.273& 0.250 & 0.278 \\
            &\multicolumn{1}{c|}{}& 720   & 0.307 & 0.324 & \textbf{0.302} & 0.322 & 0.308 & \textbf{0.321} & 0.309 & 0.326 & 0.321 & 0.328& 0.320 & 0.335& 0.322 & 0.326& 0.321 & 0.330 \\
			\cline{2-19}
			&\multirow{4}*{\rotatebox{90}{Traffic}}& 96    & \textbf{0.362} & 0.242 & 0.370 & 0.252 & 0.369 & 0.237 & 0.367 & 0.252 & 0.364 & 0.240& 0.367 & 0.251& 0.379 & \textbf{0.231}& 0.382 & 0.234 \\
            &\multicolumn{1}{c|}{} & 192   & \textbf{0.382} & 0.252 & 0.388 & 0.258 & 0.388 & 0.244 & 0.386 & 0.256 & \textbf{0.382} & 0.245 & 0.385 & 0.259& 0.398 & \textbf{0.239}& 0.403 & 0.245 \\
            &\multicolumn{1}{c|}{}& 336   & \textbf{0.392} & 0.257 & 0.400 & 0.266 & 0.398 & 0.246 & 0.400 & 0.267 & 0.396 & 0.253 & 0.398 & 0.265 & 0.411 & 0.246 & 0.419 & 0.257\\
            &\multicolumn{1}{c|}{}& 720   & \textbf{0.428} & 0.282 & 0.436 & 0.288 & 0.429 & 0.266 & 0.435 & 0.290 & 0.434 & 0.277 & 0.434 & 0.287 & 0.443 & \textbf{0.265} & 0.460 & 0.296\\
            \cline{2-19}
			&\multirow{4}*{\rotatebox{90}{Electricity}}& 96    & \textbf{0.128} & 0.221 & \textbf{0.128} & 0.221 & \textbf{0.128} & \textbf{0.217} & 0.130 & 0.224 & 0.131 & 0.223 & 0.130 & 0.222 & 0.131 & 0.224 & 0.131 & 0.223 \\
			&\multicolumn{1}{c|}{}& 192   & 0.144 & 0.237 & \textbf{0.142} & 0.236 & 0.143 & \textbf{0.233} & 0.145 & 0.240 & 0.148 & 0.239 & 0.148 & 0.240 & 0.149 & 0.241 & 0.148 & 0.240 \\
			&\multicolumn{1}{c|}{}& 336   & 0.164 & 0.257 & 0.163 & 0.255 & \textbf{0.162} & \textbf{0.252} & 0.166 & 0.260 & 0.165 & 0.256 & 0.167 & 0.261 & 0.165 & 0.257 & 0.167 & 0.257 \\
			&\multicolumn{1}{c|}{}& 720   & 0.201 & 0.290 & \textbf{0.199} & 0.289 & \textbf{0.199} & \textbf{0.284} & 0.204 & 0.293 & 0.208 & 0.293 & 0.202 & 0.291 & 0.207 & 0.290 & 0.207 & 0.290\\
			\cline{2-19}
   			&\multirow{4}*{\rotatebox{90}{ETTh1}}& 96    & 0.355 & 0.383 & 0.354 & 0.384 & \textbf{0.353} & 0.379 & 0.356 & 0.384 & 0.376 & 0.401 & 0.375 & 0.399 & 0.367 & 0.392 & 0.376 & 0.400 \\
			&\multicolumn{1}{c|}{}& 192   & \textbf{0.373} & 0.394 & 0.376 & 0.397 & 0.376 & \textbf{0.392} & 0.375 & 0.394 & 0.411 & 0.418 & 0.414 & 0.421 & 0.411 & 0.416 & 0.412 & 0.418 \\
			&\multicolumn{1}{c|}{}& 336   & \textbf{0.391} & \textbf{0.410} & 0.397 & 0.421 & 0.396 & \textbf{0.410} & 0.394 & 0.411 & 0.429 & 0.432 & 0.431 & 0.436 & 0.431 & 0.427 & 0.430 & 0.431 \\
			&\multicolumn{1}{c|}{}& 720   & 0.446 & 0.463 & 0.446 & 0.462 & \textbf{0.437} & \textbf{0.450} & 0.444 & 0.462 & 0.445 & 0.462 & 0.449 & 0.466 & 0.443 & 0.455 & 0.442 & 0.460 \\
			\cline{2-19}
   			&\multirow{4}*{\rotatebox{90}{ETTh2}}& 96    & \textbf{0.220} & 0.298 & 0.226 & 0.300 & 0.224 & \textbf{0.296} & 0.222 & 0.298 & 0.275 & 0.334 & 0.274 & 0.336 & 0.277 & 0.331 & 0.276 & 0.334 \\
			&\multicolumn{1}{c|}{}& 192   & \textbf{0.267} & 0.332 & 0.276 & 0.335 & 0.272 & \textbf{0.331} & 0.270 & 0.333 & 0.340 & 0.375 & 0.339 & 0.379 & 0.343 & 0.374 & 0.341 & 0.375 \\
			&\multicolumn{1}{c|}{}& 336   & \textbf{0.304} & \textbf{0.363} & 0.319 & 0.368 & 0.311 & 0.364 & 0.307 & 0.364 & 0.329 & 0.378 & 0.331 & 0.380 & 0.333 & 0.378 & 0.331 & 0.378 \\
			&\multicolumn{1}{c|}{}& 720   & \textbf{0.375} & 0.417 & 0.395 & 0.427 & 0.380 & \textbf{0.416} & 0.377 & 0.417 & 0.378 & 0.419 & 0.379 & 0.422 & 0.382 & 0.417 & 0.380 & 0.419 \\
			\cline{2-19}
   			&\multirow{4}*{\rotatebox{90}{ETTm1}}& 96    & 0.290 & 0.340 & 0.292 & 0.341 & 0.290 & \textbf{0.334} & \textbf{0.289} & 0.339 & 0.290 & 0.338 & 0.290 & 0.342 & 0.294 & 0.330 & 0.294 & 0.330 \\
			&\multicolumn{1}{c|}{}& 192   & \textbf{0.325} & 0.361 & 0.326 & 0.362 & 0.328 & \textbf{0.357} & 0.327 & 0.362 & 0.334 & 0.365 & 0.332 & 0.369 & 0.339 & 0.359 & 0.337 & 0.358 \\
			&\multicolumn{1}{c|}{}& 336   & \textbf{0.353} & 0.382 & 0.354 & 0.382 & 0.355 & \textbf{0.377} & 0.355 & 0.382 & 0.359 & 0.382 & 0.366 & 0.392 & 0.361 & 0.378 & 0.362 & 0.378\\
			&\multicolumn{1}{c|}{}& 720   & \textbf{0.413} & 0.413 & 0.417 & 0.413 & 0.415 & \textbf{0.409} & 0.416 & 0.413 & 0.421 & 0.420 & 0.420 & 0.424 & 0.415 & 0.414 & 0.415 & 0.414 \\
			\cline{2-19}
   			&\multirow{4}*{\rotatebox{90}{ETTm2}}& 96 & \textbf{0.164} & 0.251 & 0.168 & 0.253 & 0.165 & 0.249 & \textbf{0.164} & 0.251 & 0.165 & 0.250 & 0.165 & 0.255 & 0.164 & \textbf{0.246} & 0.164 & 0.246 \\
			&\multicolumn{1}{c|}{}& 192   & 0.220 & 0.291 & 0.224 & 0.291 & 0.219 & 0.285 & 0.219 & 0.289 & 0.219 & 0.289 & 0.220 & 0.292 & \textbf{0.215} & \textbf{0.283} & 0.218 & 0.285 \\
			&\multicolumn{1}{c|}{}& 336   & 0.264 & 0.322 & 0.265 & 0.320 & 0.265 & 0.318 & \textbf{0.261} & \textbf{0.317} & 0.275 & 0.326 & 0.278 & 0.329 & 0.270 & 0.320 & 0.270 & 0.323 \\
			&\multicolumn{1}{c|}{}& 720   & \textbf{0.342} & 0.375 & 0.343 & \textbf{0.370} & 0.347 & \textbf{0.370} & 0.345 & 0.371 & 0.365 & 0.382 & 0.367 & 0.385 & 0.355 & 0.374 & 0.363 & 0.380 \\
			\cline{2-19}
   &\multicolumn{2}{c|}{Avg.}& \textbf{0.292} & 0.316 & 0.296 & 0.318 & 0.294 & \textbf{0.312} & 0.294 & 0.318 & 0.305 & 0.322 & 0.306 & 0.327 & 0.307 & 0.318 & 0.309 & 0.322\\
			\cline{2-19}
		\end{tabular}
	}
 \end{small}
\caption{Ablation study of loss functions for training in PatchMixer. 4 cases are included: (a) both MSE and MAE are included in loss function; (b) MSE; (c) MAE; (d) SmoothL1loss. The best results are in \textbf{bold}. }
\label{tab::loss}
\end{table*}
\linespread{1}

\section{Conclusion}

In LTSF, since intra-variable mutual information in time series benchmark datasets is denser than that among variables, we hypothesize that patch mixing outperforms channel mixing in predictive performance. We validate this through PatchMixer, a CNN-based model. Our experiments demonstrate that even with models of limited receptive field, the patch-mixing architecture can improve predictive performance while accelerating training and inference. Lastly, we emphasize the importance of data optimization in LTSF tasks, which can enhance predictive performance through patch embedding and loss function optimization.

\section{Acknowledgements}

This work was supported by the National Natural Science Foundation of China (No. 62306257). The views and conclusions contained herein are those of the authors and should not be interpreted as necessarily representing the official policies or endorsements, either expressed or implied, of the National Natural Science Foundation.

\bibliography{ijcai24}
\bibliographystyle{named}

\end{document}